\PassOptionsToPackage{table}{xcolor}
\documentclass[runningheads]{llncs}
\usepackage{pdflscape}
\usepackage{graphicx}
\usepackage{multicol}
\usepackage{multirow}
\usepackage{colortbl}
\usepackage{hyperref}
\usepackage{enumitem,kantlipsum}
\usepackage{amsmath}
\usepackage{pifont}
\usepackage{lscape}
\usepackage{rotating}
\usepackage{amssymb}

\definecolor{yescolor}{cmyk}{0.66, 0, 0.35, 0.09}
\definecolor{nocolor}{cmyk}{0,0.87,0.68,0.32}
\definecolor{lowregimecolor}{cmyk}{1,0,0.07,0.6}
\definecolor{highregimecolor}{cmyk}{1,0.25,0,0.13}
\definecolor{partialcolor}{rgb}{0.302,0.302,0.302}
\definecolor{questionmarkcolor}{cmyk}{0.1,0.5,0.4,0.3}

\newlength{\maxlen}

\usepackage{etoolbox}
\patchcmd{\paragraph}{\itshape}{\bfseries\boldmath}{}{}

\begin{document}
\title{How Different Is Stereotypical Bias Across Languages?}

\titlerunning{How Different Is Stereotypical Bias Across Languages?}

%\author{First Author\inst{1}\orcidID{0000-1111-2222-3333} \and Second Author\inst{2,3}\orcidID{1111-2222-3333-4444} \and Third Author\inst{3}\orcidID{2222--3333-4444-5555}}
%\authorrunning{F. Author et al.}
%\institute{Princeton University, Princeton NJ 08544, USA \and Springer Heidelberg, Tiergartenstr. 17, 69121 Heidelberg, Germany \email{lncs@springer.com}\\ \url{http://www.springer.com/gp/computer-science/lncs} \and ABC Institute, Rupert-Karls-University Heidelberg, Heidelberg, Germany\\ \email{\{abc,lncs\}@uni-heidelberg.de}}

\author{Ibrahim Tolga Öztürk\inst{1} \and Rostislav Nedelchev\inst{2} \and Christian Heumann\inst{1} \and Esteban Garces Arias\inst{1} \and Marius Roger\inst{1} \and Bernd Bischl\inst{1,3} \and Matthias Aßenmacher\inst{1,3}}

\authorrunning{Öztürk et al.}

\institute{Department of Statistics, LMU Munich, Germany \\
\email{i.ozturktolga@gmail.com}\\
\email{\{chris,esteban.garcesarias,marius.roger,bernd.bischl,matthias\}\\@stat.uni-muenchen.de}
\and
Smart Data Analytics (SDA), University of Bonn, Germany
\email{rostislav.nedelchev@uni-bonn.de}
\and 
Munich Center for Machine Learning (MCML), LMU Munich, Germany
}

\maketitle              % typeset the header of the contribution

\begin{abstract}

Recent studies have demonstrated how to assess the stereotypical bias in pre-trained English language models. In this work, we extend this branch of research in multiple different dimensions by systematically investigating (a) mono- and multilingual models of (b) different underlying architectures with respect to their bias in (c) multiple different languages. To that end, we make use of the English Stereo\-Set data set \cite{nadeem-etal-2021-stereoset}, which we semi-automatically translate into German, French, Spanish, and Turkish. We find that it is of major importance to conduct this type of analysis in a multilingual setting, as our experiments show a much more nuanced picture as well as notable differences from the English-only analysis. The main takeaways from our analysis are that mGPT-2 (partly) shows surprising anti-stereotypical behavior across languages, English (monolingual) models exhibit the strongest bias, and the stereotypes reflected in the data set are least present in Turkish models. Finally, we release our codebase alongside the translated data sets and practical guidelines for the semi-automatic translation to encourage a further extension of our work to other languages. 

\keywords{Stereotypes \and Bias \and Fairness \and Natural Language Processing \and Pre-Trained Language Models \and Transformer \and Benchmarking}
\end{abstract}
\section{Introduction}
\label{sec:intro}

Stereotypical bias in pre-trained language models (PLMs) has been an actively researched topic in contemporary natural language processing, with the concept of \textit{gender} likely being the most prominent one among the examined demographic biases \cite{ws-2019-gender,gebnlp-2020-gender,gebnlp-2021-gender,gebnlp-2022-gender}. Since PLMs primarily learn from the data gathered from pages and websites open to and created by the public, they also inevitably memorize the stereotypes\footnote{A generalized belief about a particular category of people \cite{cardwell2014dictionary}.} present in this data. On one hand, it is infeasible to inspect individual entries one-by-one in a data set to ensure it does not possess any stereotypes, due to typically large data set sizes; on the other hand, the data set cannot be considerably downsized, as this would limit the performance of the machine learning model. 
Stereotypical decisions driven by predictions derived from deep learning models can render companies or engineers to be liable for the stereotypical bias. Hence, the likelihood of producing stereotypical outputs must be minimized, and before that, a generic methodology to measure and evaluate the stereotypical bias in the models is essential. To this day, various approaches for stereotypical bias measurement exist in the literature. An inspired approach to measure stereotypical bias in the pre-trained language models was proposed by Nadeem et al. \cite{nadeem-etal-2021-stereoset}, where an English data set and a methodology to measure the stereotypical bias in English language models was constructed. However, this methodology is significantly limited, as it supports only one language, whereas the current state-of-the-art multilingual models support more than 90 languages \cite{multilingual_lms}.

\paragraph{Contribution} In this work, we evaluate the stereotypical bias in mono- and multilingual models by creating new data sets via semi-automated translation of the StereoSet data \cite{nadeem-etal-2021-stereoset} to four different languages. This enables us to draw comparisons across multiple different dimensions and obtain a more nuanced picture. We determine to which extent pre-trained language models exhibit stereotypical biases by carefully considering multiple different combinations: We 1) examine both mono- and multilingual models, while 2) considering the different commonly used transformer architectures (encoder, decoder, encoder-decoder) and 3) perform our experiments for languages of different families (Indo-European vs. Ural-Altaic). 
In a series of experiments, we extend the code\footnote{\href{https://github.com/moinnadeem/StereoSet}{https://github.com/moinnadeem/StereoSet}} published by Nadeem et al. \cite{nadeem-etal-2021-stereoset} to a more generic version allowing for easier application to other languages and models. Additionally, we noticed and corrected some inconsistencies in this code, which we will further discuss in Section \ref{sec:multi_class}. We publish our codebase
\footnote{\href{https://github.com/slds-lmu/stereotypes-multi}{https://github.com/slds-lmu/stereotypes-multi}} 
to nurture further research with respect to stereotypical bias. 

%----------------------------------------------------------------------------------------

\section{Related work}
\label{sec:related}

Detecting and mitigating bias and stereotypes in PLMs represents an active and relevant research field, especially since these stereotypes might actually lead to negative real-world consequences for humans. Thus, it has become common practice, to at least try to measure biases and stereotypes when pre-training a new model. The word embedding association tests (WEAT \cite{bolukbasi_2016}) is one important example, showing that European-American names have more positive valence than African-American names in state-of-the-art sentiment analysis applications. Caliskan et al. \cite{Caliskan_2017} claim that this issue pertains to a much broader context than having intentional bias among different groups of people, as it is more challenging to analyze the underlying reasons for this behavior. Nadeem et al. \cite{nadeem-etal-2021-stereoset} measure the stereotypical bias (for the English language) by creating their own data set, with WEAT being the inspiration for their so-called Context Association Test (CAT). Although this (as well as most other) work is conducted on English PLMs, there is also a notable amount of research on multi-lingual models. For instance, Stanovsky et al. \cite{machine_translation_bias} conduct an experiment on the comparison of gender bias in some of the widely used translation services. They discovered that Amazon Translate performs second best in the German language among the chosen systems. Moreover, three out of four systems attain the most satisfactory performance for German among eight different languages. A rationale for that might be German's similarity to the English source language. Lauscher and Glavas \cite{lauscher_2019} measure different types of cross-lingual biases in seven languages from various language families. They come to the unanticipated finding that the Wikipedia corpus is more biased than a corpus of tweets. Further, their results indicate that FastText is the most biased method among the four examined embedding models. Névéol et al. \cite{neveol-etal-2022-french} extend the CrowS-Pairs data set \cite{nangia-etal-2020-crows} to the French language and measure the bias while providing the possibility to extend to different languages.

Other than that, there is also work on the sources of bias and on mitigation (i.e., debiasing). Mehrabi et al. \cite{bias_survey} divide the sources of bias into two categories: originating from the data and originating from the model. The behavior of a model overly focusing on data-related biases is called bias amplification \cite{Zhao_2017}. Hall et al. \cite{bias_amp} report a correlation between the strength of bias amplification and measures such as accuracy, model capacity, or model overconfidence. This also implies that this issue is more substantial when recognizing group membership (e.g., gender) is easier than class membership (e.g., positive). Besides introducing WEAT, Bolukbasi et al. \cite{bolukbasi_2016} also propose debiasing techniques. Bartl et al. \cite{bert_debias} apply counterfactual data substitution to the GAP corpus \cite{webster2018mind} and fine-tune BERT \cite{devlin-etal-2019-bert} to mitigate gender bias, achieving promising results for English. However, the same method yielded unsatisfactory performance for German -- possibly due to grammar, since German is a gender-marking language, in contrast to English. This shows once more that bias detection and mitigation depend on the language, stressing the importance of our work. Going beyond gender, Meade et al. \cite{stereoset_debias} also apply debiasing techniques for racial and religious biases.

%----------------------------------------------------------------------------------------

\section{Materials and Methods}
\label{sec:material}

\subsection{StereoSet data}
\label{sec:data}

The StereoSet data set, created by Nadeem et al. \cite{nadeem-etal-2021-stereoset}, is designed to have two association tests (intra- and inter-sentence) for the evaluation of pre-trained models. For the intra-sentence test, the model predicts the probability for the occurrence of specific words \textit{within} a sentence, which is essentially a fill-in-the-blank task. Three given candidates -- where one is deemed "stereotypical", one "anti-stereotypical", and one "unrelated" -- are inspected, and the predictions are used to calculate a score for the model.
Inter-sentence tests roughly correspond to BERT's Next Sentence Prediction (NSP) task. Again, three candidates belonging to the above-mentioned categories are considered, and the model's choice is expressed by ranking the three options. Examples for both tasks are depicted in Table \ref{table:examples}. The "unrelated" category exists to measure the general performance of the model, i.e., to check whether the model prefers a meaningful option (i.e., stereotypical or anti-stereotypical) over the unrelated option. The final score (cf. Sec. \ref{sec:eval}) thus measures the biasedness as well as the language modeling capabilities.

\begin{table}[ht]
\centering
\caption{Example of the two types of tests.}
\label{table:examples}
\resizebox{.6\textwidth}{!}{
\begin{tabular}{ |c| } 
 \hline
 \\
 \textbf{Intra-sentence task} \\ 
 \textcolor{purple}{Context Sentence:} The Muslim was extremely \_\_\_\_ in his practice. \\ 
 \textcolor{purple}{Candidate Word 1 (stereotype):} violent\\
 \textcolor{purple}{Candidate Word 2 (anti-stereotype):} non-violent\\ 
 \textcolor{purple}{Candidate Word 3 (unrelated):} orange\\ 
 \textcolor{blue}{Target:} Muslim \qquad \textcolor{blue}{Bias Type:} Religion\\\\
 \hline
 \multicolumn{1}{c}{\mbox{}}
\end{tabular}
}
\centering
\resizebox{.6\textwidth}{!}{
\begin{tabular}{ |c| } 
 \hline
 \\
 \textbf{Inter-sentence task} \\
 \textcolor{purple}{Context Sentence:} My professor is a Hispanic man. \\ 
 \textcolor{purple}{Candidate Sentence 1 (stereotype):} He came here illegally.\\ 
 \textcolor{purple}{Candidate Sentence 2 (anti-stereotype):} He is a legal citizen. \\
 \textcolor{purple}{Candidate Sentence 3 (unrelated):} The knee was bruised. \\ 
 \textcolor{blue}{Target:} Hispanic \qquad \textcolor{blue}{Bias Type:} Race\\\\
 \hline
\end{tabular}
}
\end{table}

Further, for each context sentence, the target of the stereotype (i.e., which group of people is concerned) is given. In the intra-sentence example above, the target word is "Muslim"; in the inter-sentence example, it is "Hispanic". Hence, it is possible to measure the bias for specific \textit{target groups}. Nadeem et al. \cite{nadeem-etal-2021-stereoset} used Wikidata relation triples ($<$subject, relation, object$>$) to produce these target terms, where the "relation" in these triples provides the bias type (e.g., "Gender"). Overall, there are four different \textit{bias types}: gender, profession, race, and religion. Referring again to the intra-sentence example above, the bias type is religion, while for the inter-sentence example above, it is race. The categorization is important with regard to measuring the bias per type (cf. Sec. \ref{sec:multi_class}).

Overall, there are $n = 2123$ samples in the inter-sentence\footnote{ \cite{nadeem-etal-2021-stereoset} only publish the development set, so our work is based on this.} and $n = 2106$ in the intra-sentence data set. From 79 unique target terms in the inter-sentence data set, the most common target term has 33 occurrences, and the least common has 20. For the intra-sentence data set, the target terms occur between 21 and 32 times. There are also 79 target terms for the intra-sentence test set, which makes the data set quite balanced with respect to the target terms. Regarding the bias type, there are 976 (962) examples for race, 827 (810) for profession, 242 (255) for gender, and 78 (79) for religion in the inter-sentence (intra-sentence) test sets.

%----------------------------------------------------------------------------------------

\subsection{Pre-Trained Models}
\label{sec:models}

We evaluate all three different commonly used pre-trained transformer architectures: encoder, decoder, and encoder-decoder. As a representative for the first type, we chose BERT, for the second one GPT-2 \cite{gpt2}, and for the third one T5 \cite{t5}. For each architecture, we evaluate monolingual models as well as their multilingual counterparts. While BERT was pre-trained using Masked Language Modeling (MLM) and the NSP objective, GPT-2 was on the language modeling objective. T5 relies on a pre-training objective similar to MLM but replaces entire corrupted spans instead of single tokens. Further, the English T5 models on huggingface \cite{huggingface} are already fine-tuned on 24 tasks. Appendix \ref{a:models} holds an overview of the specific models we evaluate. For Turkish, no pre-trained monolingual T5 model was available (as of the time of writing).

%----------------------------------------------------------------------------------------

\subsection{Evaluation}
\label{sec:eval}

The model predictions are not only evaluated with respect to their biasedness but also with respect to their syntactic/semantic meaningfulness. A random model that always outputs random candidates would be non-stereotypical, but it would not have any language modeling capabilities. The ideal model should excel in language modeling while simultaneously exhibiting fair behavior. Therefore, a Language Modeling Score (LMS), as well as a Stereotype Score (SS), are calculated and combined to the \textit{Idealized Context Association Test} (ICAT) score, as proposed by Nadeem et al. \cite{nadeem-etal-2021-stereoset}.\footnote{Although the work by Nadeem et al. \cite{nadeem-etal-2021-stereoset} serves as our main inspiration, there are differences regarding evaluation. See Appendix \ref{a:inter_pred_gen} and \ref{a:diff_nadeem} for the differences and our corrections.}

\paragraph{Stereotype Score (SS)} This score is designed to assess the potential amount of stereotypes in a model by comparing its preference of the stereotypical ($x_{stereo}$) over anti-stereotypical ($x_{anti}$) candidates, and vice versa.\footnote{Note that always preferring an anti-stereotypical candidate is also appraised as discriminatory behavior since it would also create unfairness towards the stereotypical group.} Thus, solely a model that prefers neither $x_{stereo}$ nor $x_{anti}$ candidates systematically is considered unbiased. SS calculation is depicted in Eq. \ref{eq:ss_score}, where a model with a score of 50\% is considered unbiased.

\begin{align}
SS &= \dfrac{1}{n}\sum_{i=1}^{n} {g(x_i)}*100, \: 
\label{eq:ss_score}\\
\text{with}\: &g(x)=
\begin{cases} 
      1, & (x_{stereo} > x_{anti}) \\  
      0, & (x_{stereo} < x_{anti})
\end{cases}\notag
\end{align}

\paragraph{Language Modelling Score (LMS)} Language modeling capabilities are assessed by measuring the number of cases in which the model prefers $x_{stereo}$ and/or $x_{anti}$ over the unrelated candidate ($x_{unr}$). The ideal model should always prefer both of them over the \texttt{unr} candidate, thus achieving an LMS of 100\%. Again, we slightly deviate from \cite{nadeem-etal-2021-stereoset}, since there are inconsistencies with their definition (cf. Appendix \ref{a:diff_nadeem}):

\begin{equation}
\begin{aligned}
LMS &= \frac{1}{2n}\sum_{i=1}^{n} {g(x_i)}*100,\: \text{with} \\
g(x) &=
\begin{cases} 
      2, & (x_{stereo} > x_{unr}) \land (x_{anti} > x_{unr}) \\
      
      1, & (x_{stereo} > x_{unr}) \land (x_{anti} < x_{unr}) \\

      1, & (x_{stereo} < x_{unr}) \land (x_{i_{anti}} > x_{unr}) \\
      
      0, & (x_{stereo} < x_{unr}) \land (x_{anti} < x_{unr})
\end{cases}
\end{aligned}
\label{eq:lm_score}
\end{equation}

\paragraph{Idealized CAT (ICAT) Score} This score combines both SS and LMS to overcome the trade-off between the two of them and allow for a holistic evaluation:

\begin{equation}
ICAT = LMS * \dfrac{min(SS, 100-SS)}{50}
\label{eq:icat_score}
\end{equation}

A completely unbiased model which always prefers meaningful candidates (i.e., $SS = 50$, $LMS = 100$) would produce an ICAT score of 100, whereas an entirely random model (i.e., $SS = 50$, $LMS = 50$) would score 50. A model that \textit{always} picks the stereotypical over the anti-stereotypical candidate (or vice versa) would result in $ICAT = 0$.

\subsection{Multi-Class Perspective}
\label{sec:multi_class}

Nadeem et al. \cite{nadeem-etal-2021-stereoset} considered the four different bias types as classes and were thus able the evaluate the models in a multi-class fashion. Nevertheless, there were some mistakes in this setting which we attempt to correct. While we define $ICAT_{macro}$ as the average over the bias type-specific $ICAT$ scores and $ICAT_{micro}$ as the calculation of the $ICAT$ over the averaged sub-scores ($LMS$ and $SS$), their definition was exactly the other way round. We were in close contact with Nadeem et al. \cite{nadeem-etal-2021-stereoset} to discuss this disagreement and they also confirmed our point of view.

%----------------------------------------------------------------------------------------

\section{Methods for Probability Predictions}
\label{sec:pred}

\subsection{Intra-Sentence Predictions}
\label{sec:intra_pred}

\begin{figure}[ht]
\centering
\caption{An example for multiple mask tokens. There are six different sentences to be processed for only one example in this case.}
\label{fig:multi_mask_token}
\resizebox{.5\textwidth}{!}{
\usetikzlibrary{automata}
\begin{tikzpicture}[sibling distance=170pt, box/.style={rectangle,draw}] 
   \node[box] {The chess player was [BLANK].}
   child {node[box] {Asian} 
   child {node[box, text width=3cm, align=center] {The chess player was [MASK].\\The chess player was a[MASK].}}}
   child {node[box] {Hispanic}
   child {node[box, text width=7cm, align=center] {
   The chess player was [MASK].\\
   The chess player was his[MASK].\\
   The chess player was Hispan[MASK].}}}
   child {node[box] {fox}
   child {node[box, text width=3cm, align=center] {The chess player was [MASK].}}}; 
\end{tikzpicture}
}
\end{figure}
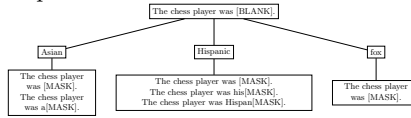

Inferring BERT and T5 for the intra-sentence tests is trivial due to their highly similar pre-training objectives described in Section \ref{sec:models}. However, GPT-2 does not have any objective related to MLM. Thus, it cannot solve this task in a discriminative manner but rather uses a generative approach. Since candidate words usually consist of multiple tokens, the probability of the whole word cannot be calculated directly. Following \cite{nadeem-etal-2021-stereoset}, the candidate word is divided into its tokens, and each token is unmasked step-by-step from left to right. After manipulating the data set this way (cf. Fig. \ref{fig:multi_mask_token}, Appendix \ref{a:intra_pred}), one sentence requires multiple inference steps. Nevertheless, due to efficient object-oriented handling, the inference can be accomplished batch-by-batch and with multiprocessing. Furthermore, instead of padding to a fixed length (as \cite{nadeem-etal-2021-stereoset}), we use dynamic padding with the aim of reducing memory consumption. After acquiring the probabilities for the masked tokens, they are averaged per candidate word.

The probability distribution for each token is generated by providing their respective left context to the model. In other words, the generation is executed for every token instead of only the masked part. Due to the left-to-right nature of the model, the masked part does not affect only one token, but also the whole context on its right. The output of this operation produces a separate distribution for each token, where each distribution expresses the likelihood of the corresponding next token. Hence, the likelihood of generating a specific token is obtained by examining the likelihood distribution output of the previous token.

In order to predict the likelihood, the model-specific BOS token is used as the left context of the first token. After calculating the likelihoods for both the first token and the whole sentence, the softmax operation is performed separately over the vocabulary dimension to flatten the results into a probability space, where each of the results is between zero and one. To merge these probabilities from each token, the following formula inspired by \cite{nadeem-etal-2021-stereoset} is used:

\begin{equation}
2^{\frac{\sum_{i=1}^{N} \log_2({P(x_i|x_0,x_1,...,x_{i-1})})}{N}},
\label{eq:generative_logs}
\end{equation}

where N is the number of tokens in the sentence.

%----------------------------------------------------------------------------------------

\subsection{Inter-Sentence Predictions}
\label{sec:inter_pred}

\paragraph{Discriminative Approach}

For BERT and mBERT, inter-sentence tests can be conducted by taking advantage of the discriminative NSP objective and using it to rank the candidate sentences. However, T5 and GPT-2 models were not pre-trained on NSP and must consequently be fine-tuned using this objective (cf. Sec. \ref{sec:nsp-fine}). An alternative approach would be to predict the probability for each word in the next sentence, making use of the generative nature of these models. We report more experimental results on the comparison of the discriminative and the generative evaluation approach in Appendix \ref{a:inter_pred_gen}. 

\paragraph{Generative Approach}

For the generative approach, the inference process (including tokenization) differs substantially between T5 and GPT-2 based models. In T5 models, candidate sentences are fully masked, although this hinders predicting the whole next sentence for the model. The general form of the input sentence to the encoder is "$<$context sentence$>$ $<$extra\_id\_0$>$". A specific example is "\textit{My professor is a Hispanic man. $<$extra\_id\_0$>$}". To handle this cumbersome prediction, we use teacher forcing with the inputs to the decoder having the form "\textit{$<$pad$>$ $<$extra\_id\_0$>$ $<$candidate sentence$>$}"; a specific example would be "$<$extra\_id\_0$>$ He is a legal citizen.". After obtaining the probabilities for each token, they are combined by again applying Equation \ref{eq:generative_logs}.

For inferring GPT-2, context and candidate sentences are merged, separated by whitespace "$<$context sentence$>$ $<$candidate sentence$>$" (called \textit{"full sentence"}). A specific example would be "\textit{My professor is a Hispanic man. He is a legal citizen.}". Nadeem et al. \cite{nadeem-etal-2021-stereoset} measure the final score by calculating the probability ratio of the candidate over the context, which does in fact not evaluate their dependence, but treats them entirely separately. Their results for this approach are not satisfying, which we suspect to be due to using a wrong ratio. We show that it is possible to achieve satisfying results using this generative approach for English (GPT-2 and mGPT-2) and German (mGPT-2).\footnote{Due to this finding, we abstain from fine-tuning any other monolingual GPT-2 model on NSP and rely solely on the (corrected) generative approach for this architecture.} For a more detailed explanation of our changes to the probability calculation, please refer to Appendix \ref{a:inter_pred_gen}.

%----------------------------------------------------------------------------------------

\section{Experiments}
\label{sec:experiments}

\subsection{Data Set Translation}
\label{sec:data_creat}

We translate StereoSet to German, French, Spanish, and Turkish using Amazon Web Service (AWS) translation services in Python (boto3). A crucial point in this process is translating the "BLANK" word in the context sentences in the intra-sentence data set. Since this word must be kept in the output, it is declared a special word, in the sense that it is not translated.\footnote{If left as a standard word, AWS performs various different (erroneous) translations depending on the target language/context.} We, therefore, make use of AWS's "custom terminology" approach by using the byte code \texttt{"en,de [endline] BLANK,BLANK"}\footnote{Or \texttt{fr}, \texttt{es}, \texttt{tur} instead of \texttt{de} for the other languages.} in Python to keep the BLANK token as is. After translation, all data sets were checked for punctuation errors and for the correct placement of the BLANK token in the different languages. We opted for these for languages since they exhibit several criteria which are deemed important:

\begin{itemize}
    \item[a)] German, French, and Spanish are among the most frequently spoken European languages.
    \item[b)] German, French, and Spanish have multiple grammatical genders, as opposed to English. German has three grammatical genders (der, die, das), while French (le, la) and Spanish have two (el, la).
    \item[c)] Turkish is a language from a different cultural background and does not have a grammatical gender (as does English).
\end{itemize}

\subsection{NSP Fine-Tuning}
\label{sec:nsp-fine}

Fill-in-the-blank tasks are naturally supported by all evaluated model types (cf. Sec. \ref{sec:intra_pred}). Thus, no specific fine-tuning is required for the intra-sentence data set. For mGPT-2 and T5, however, we follow \cite{nadeem-etal-2021-stereoset} by adding an NSP-head and fine-tuning these models.\footnote{We use the already fine-tuned English GPT-2 model from Nadeem et al. \cite{nadeem-etal-2021-stereoset} and the generative approach for the other GPT-2 models. All other training processes were carried out on a Tesla V100-SXM2-16GB GPU.} We use the Wikipedia data set in English, German, and French from the \texttt{datasets} library holding Wikipedia dumps extracted on March 1, 2022. Since there are no readily available data sets for Turkish and Spanish, we build them from the July 20, 2022, Wikipedia dump using the same library. After sentence-tokenizing and shuffling the data set, we add IDs to all sentences. This enables us to create consecutive sentence tuples as positive examples, while negative ones are created by drawing a random sentence.\footnote{Taking random sentences from a different article requires the model to differentiate between articles.} 

%----------------------------------------------------------------------------------------

\subsubsection{Multilingual GPT-2}
\label{sec:mgpt2_training}

For NSP fine-tuning of mGPT-2 \cite{tan2021msp}, we consider 110,000 Wikipedia articles ($\sim 9.5$M sentences) for English and German, which is a similar number of sentences used by \cite{nadeem-etal-2021-stereoset}. Due to hardware constraints, we train with a batch size of four while using gradient accumulation over 16 steps, yielding weight updates after every 64 examples. Following \cite{nadeem-etal-2021-stereoset}, we set the core learning rate to 5e-6 and to 1e-3 for the NSP-head. Training is carried out with half-precision (FP16) and terminated after around 1M examples since the accuracy stabilized at around 90\% and the loss converged (cf. Appendix \ref{a:mgpt2-details}).

\subsubsection{Monolingual T5 models and mT5}
\label{sec:t5_training}

We employ the T5 base models alongside their original tokenizers, which are both of comparable size to BERT. For fine-tuning the English T5 model, we add the prefix "\textit{binary classification: }" -- a unique wording in the T5 tokenizer -- to the start of each input sequence. After reaching satisfactory performance with mGPT-2 on only 22,000 articles ($\sim$ 1M samples), we use the same number here. Since T5 is much smaller than mGPT-2, more samples fit into GPU memory in each training step. Thus, we train with a batch size of 24 with three gradient accumulation steps to achieve a comparable number of examples per gradient update as for mGPT-2. After experimenting with FP16, the training is conducted with full precision, since FP16 training took longer for all T5-based models -- an observation that is also reported by other researchers \cite{platen_2022}. Since there is no separate NSP-head in T5 fine-tuning (as explained in Section \ref{sec:nsp-fine}), the learning rate is only set for the core model.\footnote{Appendix \ref{a:t5-details} holds the details on the scheduler.} Again, we reach an accuracy of roughly 90\% at the end of fine-tuning with converging loss (cf. Fig. \ref{fig:t5_train} in Appendix \ref{a:t5-details}). The accuracy does not seem to be fully converged, but again we refrain from committing to fully optimizing on this auxiliary task. 

We found that fine-tuning mT5 on NSP works with a relatively high (and stable) learning rate of 1e-4. To preserve comparability to mGPT-2 fine-tuning, the training is stopped after 25\% of the data set is processed, due to already achieving 92\% accuracy. We train with a batch size of eight and eight gradient accumulation steps. NSP fine-tuning for the monolingual German, French, and Spanish T5 models was performed in a similar fashion.

%----------------------------------------------------------------------------------------

\section{Results}
\label{sec:res}

As described in Section \ref{sec:multi_class}, we use two different evaluation techniques: in addition to evaluating a model as a whole, we also consider each target term as a class and treat the problem from a multi-class perspective. While Nadeem et al. \cite{nadeem-etal-2021-stereoset} only consider the multi-class results, we put a greater focus on the global evaluation of the models in order to draw conclusions with respect to the different languages and architectures.

%----------------------------------------------------------------------------------------

\subsection{Multilingual Models}
\label{sec:res_multi}

The lower part of Table \ref{table:intrasentence_eval} holds the evaluation results for the multilingual models in the intra-sentence setting. Regarding language modeling, mGPT-2 performs much better than mBERT and mT5 in all languages, which is also reflected in its higher ICAT scores. When comparing across languages, the multilingual models exhibit the highest stereotypical bias for Spanish and English, while mBERT appears to be less biased with regard to the models. The mGPT-2 model demonstrates a stereotypical bias for Spanish and English, while mT5 is quite biased for all languages. Overall, the strong mGPT-2 LMS performance leads to it also outperforming the other models with respect to ICAT, where we also observe a notable gap between English and German on the one hand and French, Spanish, and Turkish on the other hand.

Table \ref{table:intersentence_eval} provides inter-sentence evaluation results for all models.\footnote{As described in Section \ref{sec:nsp-fine}, there are two different approaches for evaluating GPT-2 and T5 models. For GPT-2, results for the generative approach are shown, while the T5 models are all fine-tuned on NSP.}
Regarding this test, mGPT-2 is outperformed by mBERT and mT5 by a large margin across languages with respect to LMS, which can probably be explained by the different pre-training regimes. Similarly, mGPT-2 behaves very differently from the other two models; while mBERT and mT5 are rather strongly biased, mGPT-2 seems to favor the anti-stereotypes across all languages.

The overall results calculated from the combination of both tests are displayed in Table \ref{table:overall_results}. All three different types of architectures exhibit a similar LMS performance, with the German language being the exception, since mT5 outperforms the other two models by a wide margin. According to SS, mGPT-2 shows either very fair behavior (en, tur, es) or even leans towards the anti-stereotype groups (as already observed in Tab. \ref{table:intersentence_eval}). The other two models on average \textit{always} prefer the stereotypical options, with the most stereotypical behavior for English and Spanish. With respect to the SS, the multilingual models' behavior seems to be the fairest for the Turkish language.\footnote{We suspect the employed data sets were collected to test primarily for \textit{western} stereotypes, since they were prepared by people from the United States. Hence, this might be one of the reasons for the apparent unbiasedness for Turkish. Future work requires building different data sets for different cultural groups.} The overall ICAT scores also reflect these findings. According to these scores, mGPT-2 is deemed the best model for English and Spanish due to its far better SS values. For German, the two other models are able to catch up a little to mT5, since it is the most biased model (despite having the best LMS). For Turkish, all the models not only exhibit similar SS, but also similar LMS values, and hence all have similar ICAT scores. Regarding the performance on the French data, mT5 beats its two competitors by showing a competitive LMS and exhibiting a low bias.

%----------------------------------------------------------------------------------------

\begin{table*}[ht]
\centering
\caption{Evaluation results for \textit{intra}-sentence tests on monolingual (top) and multilingual (bottom) models. Best score (separate for mono- and multilingual models) per language in bold.}
\label{table:intrasentence_eval}
\resizebox{1\textwidth}{!}{
\begin{tabular}{cccccccccccccccc} 
\toprule
 & \multicolumn{5}{c}{\textbf{LMS}} & \multicolumn{5}{c}{\textbf{SS}} & \multicolumn{5}{c}{\textbf{ICAT}}  \\ 

& en & de & tur & fr & es & en & de & tur & fr & es & en & de & tur & fr & es\\
\cmidrule(l{4pt}r{4pt}){2-6} \cmidrule(l{4pt}r{4pt}){7-11} \cmidrule(l{4pt}r{4pt}){12-16}
\textbf{BERT}   & 83.1   & 71.8   & 69.23   & 50.21   & 76.38 
                & \textbf{58.74}   & \textbf{55.44}   & \textbf{50.9}   & \textbf{47.67}   & 56.17 
                & 68.58   & 63.98   & 67.98   & 47.88   & \textbf{66.95} \\ 
\textbf{GPT-2}  & \textbf{91.14}   & \textbf{79.91}   & \textbf{73.46}   & \textbf{80.03}   & \textbf{79.11} 
                & 61.97   & 58.54   & 53.32   & 59.78   & 58.83 
                & \textbf{69.33}   & \textbf{66.27}   & \textbf{68.57}   & \textbf{64.38}   & 65.13 \\ 
\textbf{T5}     & 79.08   & 67.67   & ---     & 50.5   & 63.44 
                & 60.02   & 55.63   & ---     & 54.13   & \textbf{55.32} 
                & 63.24   & 60.04   & ---     & 46.33   & 56.69 \\ 
\midrule
\textbf{mBERT}  & 69.94   & 65.67   & 59.07   & 62.3   & 60.16 
                & \textbf{52.37}   & 49.17   & \textbf{49.95}   & 52.42   & \textbf{52.04} 
                & 66.62   & 64.58   & 59.01   & 59.28   & 57.7 \\ 
\textbf{mGPT-2} & \textbf{86.49}   & \textbf{77.03}   & \textbf{71.49}   & \textbf{66.93}   & \textbf{70.63} 
                & 55.08   & \textbf{50.21}   & 52.8   & 48.58   & 55.22 
                & \textbf{77.7}   & \textbf{76.7}   & \textbf{67.48}   & \textbf{65.02}   & \textbf{63.25} \\ 
\textbf{mT5}    & 69.87   & 73.97   & 55.7   & 55.56   & 56.77 
                & 52.52   & 54.3   & 51.28   & \textbf{50.95}   & 53.99 
                & 66.35   & 67.6   & 54.27   & 54.5   & 52.24 \\ 
\bottomrule 
\end{tabular}
}
\end{table*}

\begin{table*}[ht]
\centering
\caption{Evaluation results for \textit{inter}-sentence tests on monolingual (top) and multilingual (bottom) models. Best score (separate for mono- and multilingual models) per language in bold.}
\label{table:intersentence_eval}
\resizebox{1\textwidth}{!}{
\begin{tabular}{cccccccccccccccc} 
\toprule
 & \multicolumn{5}{c}{\textbf{LMS}} & \multicolumn{5}{c}{\textbf{SS}} & \multicolumn{5}{c}{\textbf{ICAT}}  \\ 

& en & de & tur & fr & es & en & de & tur & fr & es & en & de & tur & fr & es\\
\cmidrule(l{4pt}r{4pt}){2-6} \cmidrule(l{4pt}r{4pt}){7-11} \cmidrule(l{4pt}r{4pt}){12-16}
\textbf{BERT}   & 88.41   & 79.67   & \textbf{83.73}   & 61.02   & 41.85 
                & 60.24   & 55.77   & 54.07   & 43.62   & \textbf{49.22} 
                & 70.3   & 70.48   & \textbf{76.9}   & 53.23   & 41.2 \\ 
\textbf{GPT-2}  & 76.57   & 77.04   & 66.51   & 66.46   & 66.93 
                & \textbf{52}   & \textbf{51.72}   & \textbf{49.51}   & \textbf{50.26}   & 47.1 
                & \textbf{73.5}   & \textbf{74.39}   & 65.85   & 66.12   & 63.06 \\ 
\textbf{T5}     & \textbf{88.48}   & \textbf{84.48}   & ---     & \textbf{80.92}   & \textbf{77.01} 
                & 60.39   & 57.18   & ---     & 56.24   & 55.16 
                & 70.1   & 72.34   & ---     & \textbf{70.82}   & \textbf{69.07} \\ 
\midrule
\textbf{mBERT}  & 82.9   & 77.27   & 78.23   & 77.51   & 76.68 
                & 57.94   & 58.03   & 53.51   & 57.04   & 57.47 
                & 69.74   & 64.86   & 73.21   & 66.59   & 65.23 \\ 
\textbf{mGPT-2} & 69.78   & 67.57   & 63.82   & 68.75   & 67.38 
                & \textbf{45.6}   & 43.48   & \textbf{48.19}   & 45.03   & \textbf{44.84} 
                & 63.64   & 58.75   & 61.51   & 61.91   & 60.43 \\ 
\textbf{mT5}    & \textbf{84.62}   & \textbf{81.96}   & \textbf{79.06}   & \textbf{82.31}  & \textbf{82.9} 
                & 58.08   & \textbf{54.83}   & 52.43   & \textbf{54.92}   & 56.67 
                & \textbf{70.9}5   & \textbf{74.05}   & \textbf{75.23}   & \textbf{74.21}   & \textbf{71.85} \\ 
\bottomrule 
\end{tabular}
}
\end{table*}

%----------------------------------------------------------------------------------------

\begin{table*}[ht]
\centering
\caption{Overall evaluation results for monolingual (top) and multilingual (bottom) models.}
\label{table:overall_results}
\resizebox{1\textwidth}{!}{
\begin{tabular}{cccccccccccccccc} 
\toprule
 & \multicolumn{5}{c}{\textbf{LMS}} & \multicolumn{5}{c}{\textbf{SS}} & \multicolumn{5}{c}{\textbf{ICAT}}  \\ 

& en & de & tur & fr & es & en & de & tur & fr & es & en & de & tur & fr & es\\
\cmidrule(l{4pt}r{4pt}){2-6} \cmidrule(l{4pt}r{4pt}){7-11} \cmidrule(l{4pt}r{4pt}){12-16}
\textbf{BERT}   & \textbf{85.76}   & 75.76   & \textbf{76.51}   & 55.64   & 59.04 
                & 59.49   & 55.61   & 52.49   & \textbf{45.64}   & \textbf{52.68} 
                & 69.48   & 67.26   & \textbf{72.69}   & 50.78   & 55.88 \\ 
\textbf{GPT-2}  & 83.83   & \textbf{78.47}   & 69.97   & \textbf{73.22}   & \textbf{73.00} 
                & \textbf{56.96}   & \textbf{55.11}   & \textbf{51.41}   & 55.00   & 52.94 
                & \textbf{72.15}   & \textbf{70.45}   & 68.00   & \textbf{65.9}   & \textbf{68.7} \\ 
\textbf{T5}     & 83.8   & 76.11   & ---     & 65.77   & 70.25 
                & 60.18   & 56.41   & ---     & 55.19   & 55.24 
                & 66.75   & 66.35   & ---     & 58.94   & 62.89 \\ 
\midrule
\textbf{mBERT}  & 76.45   & 71.5   & \textbf{68.94}   & \textbf{69.93}   & 68.46 
                & 55.17   & 53.62   & 51.74   & 54.74   & 54.76 
                & 68.55   & 66.32   & 66.54   & 63.3   & 61.93 \\ 
\textbf{mGPT-2} & \textbf{78.1}   & 72.28   & 67.64   & 67.84   & 69.00 
                & \textbf{50.32}   & \textbf{46.83}   & \textbf{50.48}   & 46.8   & \textbf{50.01} 
                & \textbf{77.6}   & 67.7   & \textbf{66.98}   & 63.49   & \textbf{68.98} \\ 
\textbf{mT5}    & 77.28   & \textbf{77.98}   & 67.43   & 68.99   & \textbf{69.89} 
                & 55.31   & 54.57   & 51.86   & \textbf{52.94}   & 55.33 
                & 69.07   & \textbf{70.86}   & 64.92   & \textbf{64.93}   & 62.43 \\ 
\bottomrule 
\end{tabular}
}
\end{table*} 

%----------------------------------------------------------------------------------------

%----------------------------------------------------------------------------------------

\subsection{Monolingual Models}
\label{sec:res_mono}

The upper parts in Tables \ref{table:intrasentence_eval}, \ref{table:intersentence_eval} and \ref{table:overall_results} show performances for different monolingual models in each column. The most striking (and possibly least surprising) finding is that the monolingual English models exhibit the best LMS across all tables, except for GPT-2\footnote{Note that the monolingual models were not fine-tuned on NSP, but use the generative approach.} on the inter-sentence test. Similar to the multilingual setting, GPT-2 models stand out in intra-sentence LMS across languages, while they struggle in inter-sentence LMS. This leads to a more balanced overall LMS performance across models, except for BERT, which severely struggles in French and Spanish. Overall, LMS performance of most monolingual models on both tests is better compared to the multilingual ones (again, except for BERT in French and Spanish).

Regarding the biasedness of the different models, we observe that English models have the most severe stereotypical tendency; each of the three English models displays more stereotypical bias than \textit{any} of the other models for \textit{any} other language. Consequently, the higher LMS performance of these models comes at a price. Comparing the different architectures, GPT-2 models appear to be least biased on the inter-sentence test, while for the intra-sentence examples and overall, all the architectures exhibit stronger biases than their multilingual counterparts.

Focusing on ICAT scores, monolingual BERT and GPT-2 models outperform the multilingual versions on the inter-sentence test (except for French and Spanish BERT models), while monolingual T5 models are a bit worse. On the intra-sentence test, the picture is more nuanced: Spanish and Turkish models are better than the multilingual ones, while the performance is mixed for English and French, and German models are always worse than their multilingual counterparts. Overall, we also observe a strong performance of the multilingual models, mostly driven by the fact that they are less stereotypically biased. The strong performance for the Turkish monolingual models is noteworthy, since they are equally less biased but stronger in LMS than the multilingual models.

%----------------------------------------------------------------------------------------

\begin{table*}[ht]
\centering
\caption{Overall multi-class evaluation results on monolingual (top) and multilingual (bottom) models. LMS and SS are averaged across the different classes.}
\label{table:mc_overall_evals}
\resizebox{1\textwidth}{!}{
\begin{tabular}{cccccccccccccccc} 
\toprule
 & \multicolumn{5}{c}{\textbf{Avg. LMS}} & \multicolumn{5}{c}{\textbf{Avg. SS}} & \multicolumn{5}{c}{\textbf{ICAT (Macro / Micro)}}  \\ 

& en & de & tur & fr & es & en & de & tur & fr & es & en & de & tur & fr & es\\
\cmidrule(l{4pt}r{4pt}){2-6} \cmidrule(l{4pt}r{4pt}){7-11} \cmidrule(l{4pt}r{4pt}){12-16}
\textbf{BERT}   & \textbf{85.77}   & 75.77   & \textbf{76.41}   & 55.67   & 58.98 
                & 59.53   & 55.59   & 52.63   & \textbf{45.77}   & \textbf{52.66} 
                & (68.17/69.42)   & (64.19/67.3)   & \textbf{(65.6/72.4)}   & (47.62/50.96)  & (52.5/55.84) \\ 
\textbf{GPT-2}  & 83.76   & \textbf{78.39}   & 69.88   & \textbf{73.18}   & \textbf{72.91} 
                & \textbf{57.00}   & \textbf{55.05}   & \textbf{51.41}   & 54.98   & 52.9 
                & \textbf{(70.22/72.03)}   & \textbf{(66.12/70.48)}   & (60.89/67.91)   & \textbf{(63.02/65.89)}  & \textbf{(64.37/68.69)} \\ 
\textbf{T5}     & 83.8   & 76.12   & ---     & 65.81   & 70.21 
                & 60.28   & 56.31   & ---     & 55.2   & 55.19 
                & (65.59/66.57)   & (62.98/66.51)   & ---   & (56.65/58.96)  & (59.02/62.93) \\ 
\midrule
\textbf{mBERT}  & 76.52   & 71.53   & \textbf{68.99}   & \textbf{69.92}   & 68.44 
                & 55.19   & 53.63   & 51.88   & 54.86   & 54.85
                & (64.64/68.58)   & (61.71/66.33)   & (59.84/66.39)   & (59.69/63.12)  & (57.89/61.8) \\ 
\textbf{mGPT-2} & \textbf{78.12}   & 72.25   & 67.58   & 67.7   & 69.04 
                & \textbf{50.43}   & \textbf{46.85}   & \textbf{50.41}   & 46.94   & \textbf{50.01} 
                & \textbf{(68.49/77.44)}   & (62.27/67.71)   & \textbf{(59.86/67.03)}   & (57.17/63.57)  & \textbf{(60.93/69.02)} \\ 
\textbf{mT5}    & 77.29   & \textbf{77.97}   & 67.38   & 68.97   & \textbf{69.92}
                & 55.33   & 54.57   & 51.96   & \textbf{52.96}   & 55.44 
                & (65.65/69.05)   & \textbf{(65.34/70.85)}   & (59.1/64.74)   & \textbf{(60.41/64.9)}  & (59.11/62.32) \\ 
\bottomrule 
\end{tabular}
}
\end{table*}

%----------------------------------------------------------------------------------------

\subsection{Multi-Class Results}
\label{sec:res_multiclass}

Assuming that the target terms constitute separate classes, most of our findings from the above sections still hold. Thus, we only report the striking the differences for the overall results in the main paper (cf. Tab. \ref{table:mc_overall_evals}) to avoid repetition.\footnote{The results for the intra-sentence tests (cf. Tab. \ref{table:mc_intrasentence_eval}) and the inter-sentence tests (cf. Tab. \ref{table:mc_intersentence_eval}) can be found in Appendix \ref{a:mc_eval}.} The multi-class perspective comes with two separate scores: a macro and a micro version of the ICAT (cf. Sec. \ref{sec:multi_class}).
The result that the macro ICAT score is consistently lower than the micro ICAT score (across all models and languages) can be explained by larger variations of the ICAT scores between the different classes. The most important takeaway from this observation is that the scores in the underrepresented classes (gender and religion) seem to be worse than for the larger classes (race and profession), since they receive disproportionately high weights in the macro ICAT.

%----------------------------------------------------------------------------------------

\section{Discussion and Future Work}
\label{sec:disc}

Probably one of the most important issues that until now has not been tackled in a holistic manner is the matter of how to take into account the differences in stereotypes in different cultural groups. For the Turkish language, we observe consistently lower measurements of stereotypical bias in the models, which we suspect to potentially originate from cultural differences. Furthermore, we did not address differences between different models of the same architecture within languages. This is also an important endeavor for the future since it allows for comparisons of the biasedness of different pre-training regimes. A holistic analysis -- e.g., in a similar fashion to how Choshen et al. \cite{choshen2022start} execute analyses for model performance across tasks -- is necessary for advancing applied research in this direction.
Another undeniable shortcoming of current research with respect to the stereotypical behavior of PLMs is that there is a variety of different (English) data sets covering different aspects, but no holistic (multilingual) framework. Efforts in the direction of building something similar to what Ribeiro et al. \cite{ribeiro-etal-2020-beyond} created for behavioral testing might be a promising goal to move forward towards. This might even become more compelling when evaluating models like the recently introduced ChatGPT \cite{chatgpt}.

To conclude, we provide a blueprint for the assessment of stereotypical bias in a multilingual setting, which is easily extendable to other models and languages. Our analysis reveals insights into the differences between the different languages and architectures when evaluated with these data sets. The overall picture drawn by this analysis is, admittedly, quite heterogeneous and does not allow drawing a conclusion declaring one architecture the clear winner. Weighting both scores (LMS and SS) equally gives them similar importance, which might also be a debatable choice depending on the intended use case of the model. Taking this into account, we would argue that it is rather up to the user to decide on the preferable model by considering all aspects of the respective application. Thus, we believe, that our results can nevertheless be used as meaningful starting points for drawing tentative conclusions or for generating new research questions in this domain. 

\section*{Ethics statement}

\paragraph{Limitations} Most certainly, analyses like ours do not come without debatable aspects, especially when it comes to the creation as well as the translation of the employed samples. Working on this set of four bias types is non-exhaustive and should definitely be extended and refined in the future. Furthermore, translating sentences from a language with two grammatical genders to languages with three genders also comes with shortcomings, since certain grammatical constructions favor specific (anti-)stereotypical candidates in the data sets. This issue appeared to be most striking for the French language. During our semi-automated translations, we also noticed errors in the original English data sets. Still, we decided for the moment to take them as is to keep our work comparable to \cite{nadeem-etal-2021-stereoset}. For future work, we plan to carefully re-evaluate all the data sets manually. A proceeding for this might be to have native speakers of each target language check and correct every sentence of their translation of the respective data set for semantic and stylistic errors. However, this would both defeat the purpose of having it translated automatically and necessitate greater manpower than is currently available, roughly corresponding to creating the data set from scratch.

With respect to model size, our analysis is restricted to PLMs of small to medium size. Therefore it is not necessarily valid to transfer the findings to larger models, like e.g., the largest models of the GPT or T5 family. Regarding the computational requirements of our study it is important to note that assessing GPT-2 models is cheap, since the generative approach works well, whereas, for T5 models, NSP fine-tuning is recommended for the inter-sentence tests.

\paragraph{Ethical considerations} When dealing with the concept of stereotypical bias, the question of ethical implications naturally arises. Utilizing crowd workers for annotating such data might expose such people to disturbing pieces of text. Given these considerations, our approach of semi-automatically translating the data is a step in the right direction. But still, we had to manually check the sentences afterward which does not reduce the exposure. Further, it is important that such a manifold and diverse, sometimes very subtle, concept of stereotypical bias is hard to grasp in an exhaustive manner. As such, many more experiments and also more elaborated data sets, dealing with the matter on an even more granular level, are required in future research. Finally, it is important to state that making applications driven by large language models (e.g. ChatGPT \cite{chatgpt}) safe for public use is one of the most important requirements before they can be made available to a broader audience. As stereotypical bias is different in different languages and different cultural background, focusing only on the English language here is no real alternative.

%----------------------------------------------------------------------------------------

\section*{Acknowledgements}

This work has been partially funded by the Deutsche Forschungsgemeinschaft (DFG, German Research Foundation) as part of BERD@NFDI - grant number 460037581. It has also been partially funded by the OpenGPT-X project (BMWK 68GX21007C) in cooperation with Alexander Thamm GmbH.

% ---- Bibliography ----

\bibliography{custom}

\begin{thebibliography}{10}
\providecommand{\url}[1]{\texttt{#1}}
\providecommand{\urlprefix}{URL }
\providecommand{\doi}[1]{https://doi.org/#1}

\bibitem{bert_debias}
Bartl, M., Nissim, M., Gatt, A.: Unmasking contextual stereotypes: Measuring
  and mitigating bert's gender bias (2020). \doi{10.48550/ARXIV.2010.14534},
  \url{https://arxiv.org/abs/2010.14534}

\bibitem{bolukbasi_2016}
Bolukbasi, T., Chang, K.W., Zou, J., Saligrama, V., Kalai, A.: Man is to
  computer programmer as woman is to homemaker? debiasing word embeddings
  (2016). \doi{10.48550/ARXIV.1607.06520},
  \url{https://arxiv.org/abs/1607.06520}

\bibitem{Caliskan_2017}
Caliskan, A., Bryson, J.J., Narayanan, A.: Semantics derived automatically from
  language corpora contain human-like biases. Science  \textbf{356}(6334),
  183--186 (apr 2017). \doi{10.1126/science.aal4230},
  \url{https://doi.org/10.1126%2Fscience.aal4230}

\bibitem{cardwell2014dictionary}
Cardwell, M.: Dictionary of psychology. Routledge (2014)

\bibitem{choshen2022start}
Choshen, L., Venezian, E., Don-Yehia, S., Slonim, N., Katz, Y.: Where to start?
  analyzing the potential value of intermediate models. arXiv preprint
  arXiv:2211.00107  (2022)

\bibitem{gebnlp-2021-gender}
Costa-jussa, M., Gonen, H., Hardmeier, C., Webster, K. (eds.): Proceedings of
  the 3rd Workshop on Gender Bias in Natural Language Processing. Association
  for Computational Linguistics, Online (Aug 2021),
  \url{https://aclanthology.org/2021.gebnlp-1.0}

\bibitem{ws-2019-gender}
Costa-juss{\`a}, M.R., Hardmeier, C., Radford, W., Webster, K. (eds.):
  Proceedings of the First Workshop on Gender Bias in Natural Language
  Processing. Association for Computational Linguistics, Florence, Italy (Aug
  2019), \url{https://aclanthology.org/W19-3800}

\bibitem{gebnlp-2020-gender}
Costa-juss{\`a}, M.R., Hardmeier, C., Radford, W., Webster, K. (eds.):
  Proceedings of the Second Workshop on Gender Bias in Natural Language
  Processing. Association for Computational Linguistics, Barcelona, Spain
  (Online) (Dec 2020), \url{https://aclanthology.org/2020.gebnlp-1.0}

\bibitem{devlin-etal-2019-bert}
Devlin, J., Chang, M.W., Lee, K., Toutanova, K.: {BERT}: Pre-training of deep
  bidirectional transformers for language understanding. In: Proceedings of the
  2019 Conference of the North {A}merican Chapter of the Association for
  Computational Linguistics: Human Language Technologies, Volume 1 (Long and
  Short Papers). pp. 4171--4186. Association for Computational Linguistics,
  Minneapolis, Minnesota (Jun 2019). \doi{10.18653/v1/N19-1423},
  \url{https://aclanthology.org/N19-1423}

\bibitem{multilingual_lms}
Doddapaneni, S., Ramesh, G., Khapra, M.M., Kunchukuttan, A., Kumar, P.: A
  primer on pretrained multilingual language models (2021).
  \doi{10.48550/ARXIV.2107.00676}, \url{https://arxiv.org/abs/2107.00676}

\bibitem{bias_amp}
Hall, M., van~der Maaten, L., Gustafson, L., Adcock, A.: A systematic study of
  bias amplification (2022). \doi{10.48550/ARXIV.2201.11706},
  \url{https://arxiv.org/abs/2201.11706}

\bibitem{gebnlp-2022-gender}
Hardmeier, C., Basta, C., Costa-juss{\`a}, M.R., Stanovsky, G., Gonen, H.
  (eds.): Proceedings of the 4th Workshop on Gender Bias in Natural Language
  Processing (GeBNLP). Association for Computational Linguistics, Seattle,
  Washington (Jul 2022), \url{https://aclanthology.org/2022.gebnlp-1.0}

\bibitem{lauscher_2019}
Lauscher, A., Glavaš, G.: Are we consistently biased? multidimensional
  analysis of biases in distributional word vectors (2019).
  \doi{10.48550/ARXIV.1904.11783}, \url{https://arxiv.org/abs/1904.11783}

\bibitem{loshchilov2018decoupled}
Loshchilov, I., Hutter, F.: Decoupled weight decay regularization. In:
  International Conference on Learning Representations (2019),
  \url{https://openreview.net/forum?id=Bkg6RiCqY7}

\bibitem{stereoset_debias}
Meade, N., Poole-Dayan, E., Reddy, S.: An empirical survey of the effectiveness
  of debiasing techniques for pre-trained language models (2021).
  \doi{10.48550/ARXIV.2110.08527}, \url{https://arxiv.org/abs/2110.08527}

\bibitem{bias_survey}
Mehrabi, N., Morstatter, F., Saxena, N., Lerman, K., Galstyan, A.: A survey on
  bias and fairness in machine learning (2019).
  \doi{10.48550/ARXIV.1908.09635}, \url{https://arxiv.org/abs/1908.09635}

\bibitem{nadeem-etal-2021-stereoset}
Nadeem, M., Bethke, A., Reddy, S.: {S}tereo{S}et: Measuring stereotypical bias
  in pretrained language models. In: Proceedings of the 59th Annual Meeting of
  the Association for Computational Linguistics and the 11th International
  Joint Conference on Natural Language Processing (Volume 1: Long Papers). pp.
  5356--5371. Association for Computational Linguistics, Online (Aug 2021).
  \doi{10.18653/v1/2021.acl-long.416},
  \url{https://aclanthology.org/2021.acl-long.416}

\bibitem{nangia-etal-2020-crows}
Nangia, N., Vania, C., Bhalerao, R., Bowman, S.R.: {C}row{S}-pairs: A challenge
  dataset for measuring social biases in masked language models. In:
  Proceedings of the 2020 Conference on Empirical Methods in Natural Language
  Processing (EMNLP). pp. 1953--1967. Association for Computational
  Linguistics, Online (Nov 2020). \doi{10.18653/v1/2020.emnlp-main.154},
  \url{https://aclanthology.org/2020.emnlp-main.154}

\bibitem{neveol-etal-2022-french}
N{\'e}v{\'e}ol, A., Dupont, Y., Bezan{\c{c}}on, J., Fort, K.: {F}rench
  {C}row{S}-pairs: Extending a challenge dataset for measuring social bias in
  masked language models to a language other than {E}nglish. In: Proceedings of
  the 60th Annual Meeting of the Association for Computational Linguistics
  (Volume 1: Long Papers). pp. 8521--8531. Association for Computational
  Linguistics, Dublin, Ireland (May 2022). \doi{10.18653/v1/2022.acl-long.583},
  \url{https://aclanthology.org/2022.acl-long.583}

\bibitem{chatgpt}
{OpenAI}: Chatgpt: Optimizing language models for dialogue (2022),
  \url{https://openai.com/blog/chatgpt/}, accessed: 2023-01-10

\bibitem{platen_2022}
Platen, P.v.: Training with fp16 precision gives nan in longt5 · issue \#17978
  · huggingface/transformers (Jul 2022),
  \url{https://github.com/huggingface/transformers/issues/17978#issuecomment-1173761651}

\bibitem{gpt2}
Radford, A., Wu, J., Child, R., Luan, D., Amodei, D., Sutskever, I.: Language
  models are unsupervised multitask learners. {OpenAi Blog}  (2019)

\bibitem{t5}
Raffel, C., Shazeer, N., Roberts, A., Lee, K., Narang, S., Matena, M., Zhou,
  Y., Li, W., Liu, P.J.: Exploring the limits of transfer learning with a
  unified text-to-text transformer (2019). \doi{10.48550/ARXIV.1910.10683},
  \url{https://arxiv.org/abs/1910.10683}

\bibitem{ribeiro-etal-2020-beyond}
Ribeiro, M.T., Wu, T., Guestrin, C., Singh, S.: Beyond accuracy: Behavioral
  testing of {NLP} models with {C}heck{L}ist. In: Proceedings of the 58th
  Annual Meeting of the Association for Computational Linguistics. pp.
  4902--4912. Association for Computational Linguistics, Online (Jul 2020).
  \doi{10.18653/v1/2020.acl-main.442},
  \url{https://aclanthology.org/2020.acl-main.442}

\bibitem{machine_translation_bias}
Stanovsky, G., Smith, N.A., Zettlemoyer, L.: Evaluating gender bias in machine
  translation (2019). \doi{10.48550/ARXIV.1906.00591},
  \url{https://arxiv.org/abs/1906.00591}

\bibitem{tan2021msp}
Tan, Z., Zhang, X., Wang, S., Liu, Y.: Msp: Multi-stage prompting for making
  pre-trained language models better translators (2021)

\bibitem{webster2018mind}
Webster, K., Recasens, M., Axelrod, V., Baldridge, J.: Mind the gap: A balanced
  corpus of gendered ambiguous pronouns. Transactions of the Association for
  Computational Linguistics  \textbf{6},  605--617 (2018)

\bibitem{huggingface}
Wolf, T., Debut, L., Sanh, V., Chaumond, J., Delangue, C., Moi, A., Cistac, P.,
  Rault, T., Louf, R., Funtowicz, M., Davison, J., Shleifer, S., von Platen,
  P., Ma, C., Jernite, Y., Plu, J., Xu, C., Scao, T.L., Gugger, S., Drame, M.,
  Lhoest, Q., Rush, A.M.: Huggingface's transformers: State-of-the-art natural
  language processing (2019). \doi{10.48550/ARXIV.1910.03771},
  \url{https://arxiv.org/abs/1910.03771}

\bibitem{Zhao_2017}
Zhao, J., Wang, T., Yatskar, M., Ordonez, V., Chang, K.W.: Men also like
  shopping: Reducing gender bias amplification using corpus-level constraints
  (2017). \doi{10.48550/ARXIV.1707.09457},
  \url{https://arxiv.org/abs/1707.09457}

\end{thebibliography}
\bibliographystyle{splncs04}

\clearpage

%----------------------------------------------------------------------------------------

\appendix

%----------------------------------------------------------------------------------------

\section{Visualization of mGPT-2 NSP Fine-Tuning}
\label{a:mgpt2-details}

\begin{figure}[ht]
  \centering
  \caption{(Smoothed) Graphs for accuracy and loss for mGPT-2 NSP fine-tuning using the AdamW optimizer \cite{loshchilov2018decoupled}.}
  \label{fig:mGPT-2_progress}
  \includegraphics[width=\textwidth]{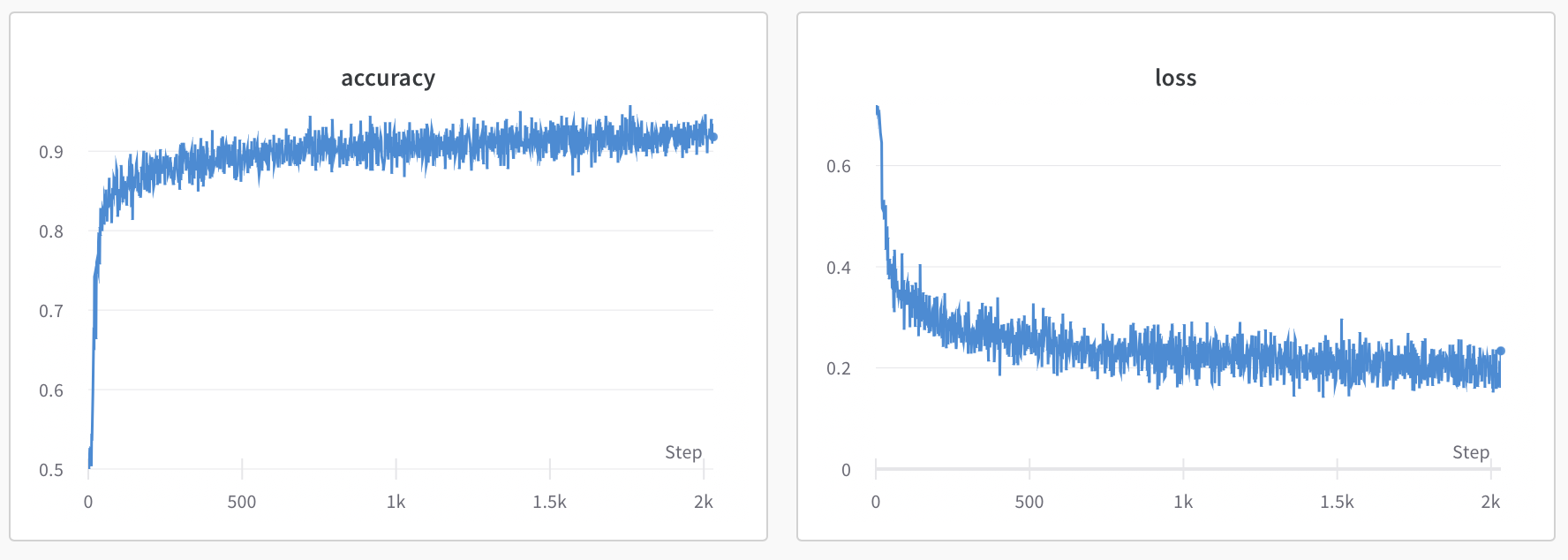}
  \includegraphics[width=\textwidth]{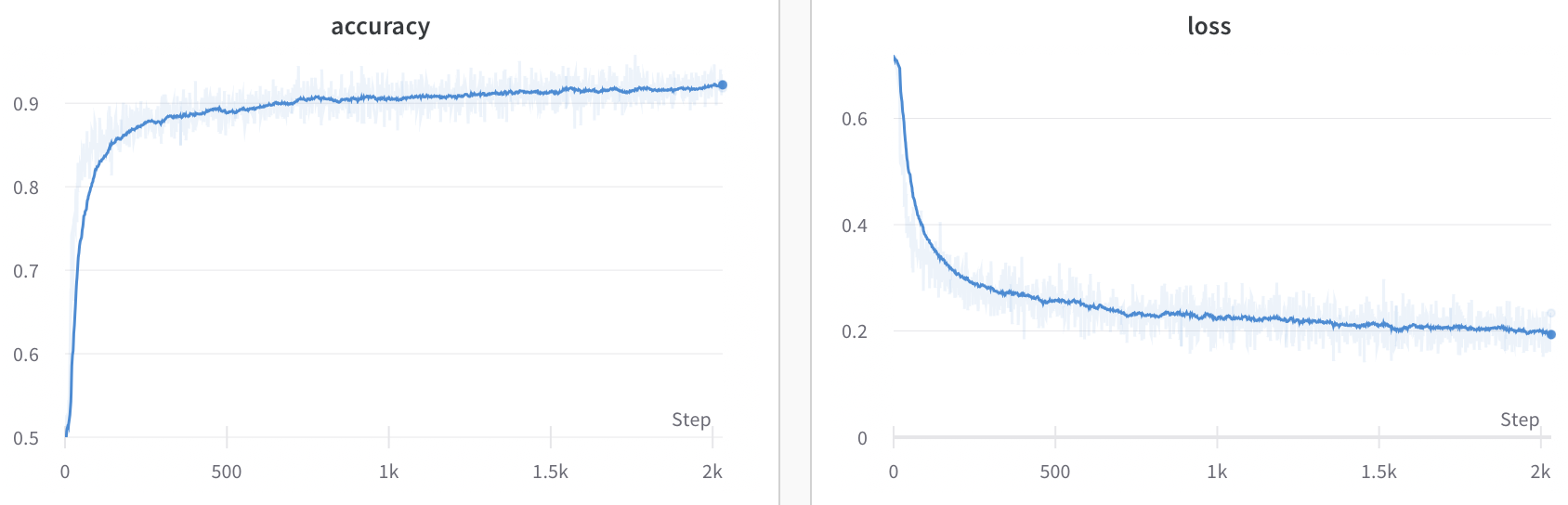}
\end{figure}

\clearpage

%----------------------------------------------------------------------------------------

\section{Visualization of T5 NSP Fine-Tuning}
\label{a:t5-details}

\begin{figure}[ht]
  \centering
  \caption{(Smoothed) Graphs for accuracy and loss for T5 NSP fine-tuning using the AdamW optimizer \cite{loshchilov2018decoupled}, as well as for the learning rate scheduler.}
  \label{fig:t5_train}
  \includegraphics[width=\textwidth]{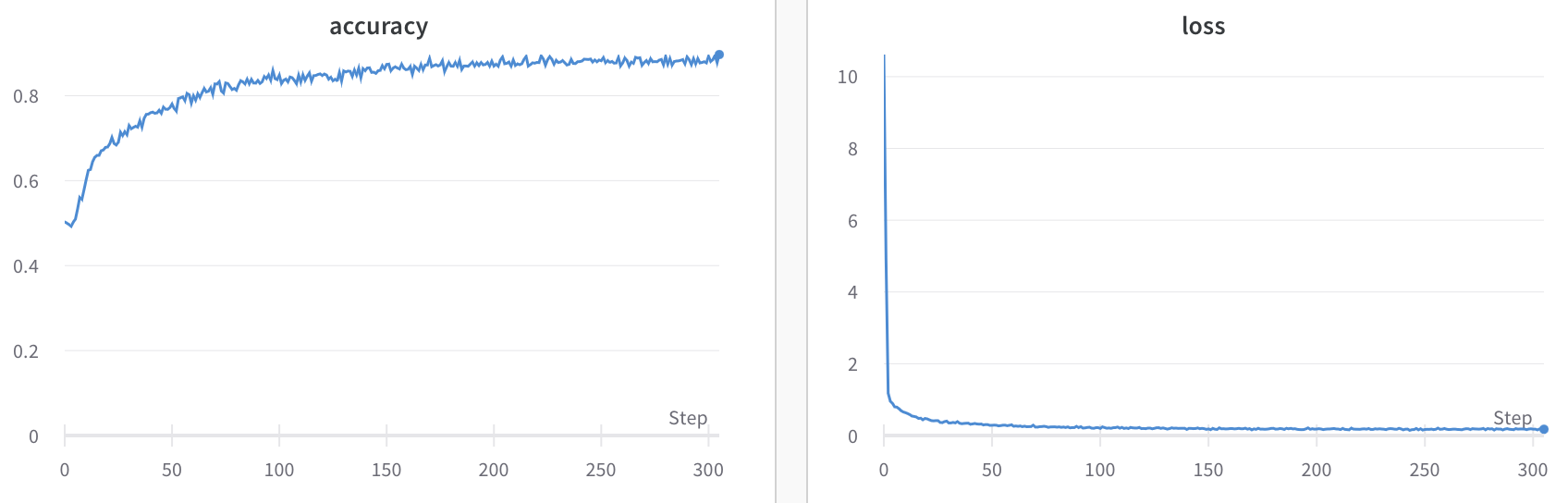}
  \includegraphics[width=\textwidth]{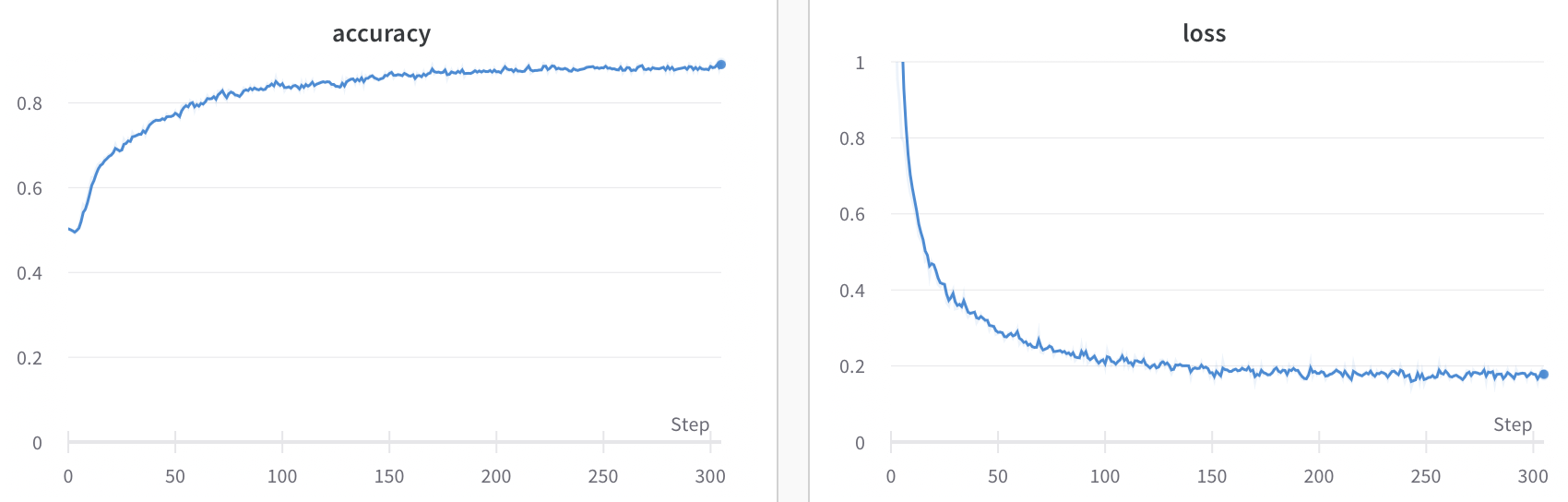}
  \includegraphics[width=.65\textwidth]{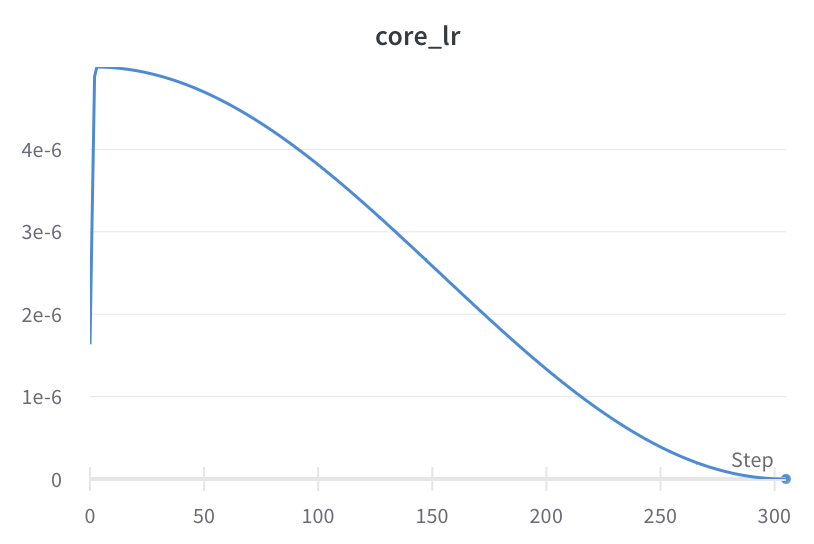}
\end{figure}

\clearpage

%----------------------------------------------------------------------------------------

\section{Model Specifications}
\label{a:models}

\begin{table}[ht]
\centering
\caption{Overview of the evaluated model architectures from huggingface. For Turkish, no pre-trained monolingual T5 model was available (as of the time of writing).}
\label{table:models}
\resizebox{.9\textwidth}{!}{
\begin{tabular}{cccc} 
\toprule
& \textbf{BERT} & \textbf{GPT-2} & \textbf{T5} \\&&&\\
\multirow{2}{*}{\textbf{multi}}   & \texttt{bert-base-}             &\multirow{2}{*}{\texttt{THUMT/mGPT}}  &  \multirow{2}{*}{\texttt{google/mt5-base}}  \\ 
								 & \texttt{multilingual-cased}      & \texttt{}                           & \texttt{} \\ &&&\\
\multirow{2}{*}{\textbf{en}}     &  \texttt{bert-base-}  & \multirow{2}{*}{\texttt{gpt-2}}   & \multirow{2}{*}{\texttt{t5-base}}  \\ 
								 & \texttt{cased}        & \texttt{}                        & \texttt{}  \\ &&&\\
\multirow{2}{*}{\textbf{de}}    & \texttt{bert-base-} & \multirow{2}{*}{\texttt{dbmdz/german-gpt2}}   & \texttt{GermanT5/t5-efficient-}  \\
								& \texttt{german-cased}   & \texttt{}   & \texttt{gc4-german-base-nl36}  \\ &&&\\
\multirow{2}{*}{\textbf{tur}}   & \texttt{dbmdz/bert-base-}   & \texttt{redrussianarmy/}   & \multirow{2}{*}{\texttt{-----}} \\ 
								& \texttt{turkish-cased}   & \texttt{gpt2-turkish-cased}   &  \texttt{}  \\ &&&\\
\multirow{2}{*}{\textbf{fr}}     & \multirow{2}{*}{\texttt{flaubert\_base\_cased}}   & \texttt{asi/gpt-fr-}   & \texttt{plguillou/t5-base-} \\ 
								& \texttt{}   & \texttt{cased-small}   & \texttt{fr-sum-cnndm}  \\ &&&\\
\multirow{2}{*}{\textbf{es}}      & \texttt{dccuchile/bert-base-}   & \texttt{PlanTL-GOB-ES/}   & \texttt{flax-community/}  \\
								& \texttt{spanish-wwm-cased}   & \texttt{gpt2-base-bne}   & \texttt{spanish-t5-small} \\ 
\bottomrule 
\end{tabular}
}
\end{table}

%----------------------------------------------------------------------------------------

\section{Data Preparation for Intra-Sentence Tests}
\label{a:intra_pred}

%----------------------------------------------------------------------------------------

\section{Generative Approach for Inter-Sentence Predictions}
\label{a:inter_pred_gen}

The score calculation approach from \cite{nadeem-etal-2021-stereoset} will be abbreviated as "\textit{gen\_orig}", while our approach, mathematically expressed as

\begin{equation}
P(cand \mid cont) = \frac{P(cand \cap cont)}{P(cont)}\;,
\label{eq:bayesian_inf}
\end{equation}

will simply be abbreviated as "\textit{gen}". In Eq. \ref{eq:bayesian_inf} $P(cont)$ is the (isolated) probability of context sentence, which can be ignored since it is the same for all candidates. Thus, the primary focus is on $P(cand \cap cont)$, which is the probability of the "full sentence". This can be measured with the probabilities of candidate sentence tokens, which are computed by considering the context sentence as their left context. Hence, this methodology implicitly contains the relationship between context sentence and candidate sentences, contrary to the work shown in \cite{nadeem-etal-2021-stereoset}. Finally, these tokens' probabilities are combined by utilizing Eq. \ref{eq:generative_logs}.

Table \ref{table:eval_gen} holds the results for the generative approach in the inter-sentence tests for English and German models:

\begin{table}[ht]
\centering
\caption{Evaluation results comparing the generative to the discriminative approach for \textit{inter}-sentence tests on monolingual (top) and multilingual (bottom) GPT-2 and T5 models for German end English.}
\label{table:eval_gen}
\resizebox{.75\textwidth}{!}{
\begin{tabular}{cccccccccccccccc} 
\toprule
 & \multicolumn{2}{c}{\textbf{LMS}} & \multicolumn{2}{c}{\textbf{SS}} & \multicolumn{2}{c}{\textbf{ICAT}}  \\ 

& en & de & en & de & en & de \\
\cmidrule(l{4pt}r{4pt}){2-3} \cmidrule(l{4pt}r{4pt}){4-5} \cmidrule(l{4pt}r{4pt}){6-7}
\textbf{GPT-2 (NSP)}   & 76.17 & ---
                & 51.91 & ---
                & 73.26 & --- \\ 
\textbf{GPT-2 (gen)}  & 76.57 & ---
                & 52   & ---
                & 73.5  & --- \\ 
\textbf{GPT-2 (gen\_orig)}  & 58.27 & ---
                & 46.21 & ---
                & 53.85 & --- \\ 
\midrule
\textbf{T5 (NSP)}     & 88.48   & ---
                & 60.39   & ---
                & 70.1   & --- \\ 
\textbf{T5 (gen)}     & 54.78   & ---
                & 54.03   & ---
                & 50.37   & --- \\ 
\midrule\midrule
\textbf{mGPT-2 (NSP)}   & 81.56   & 77.39   
                & 54.26   & 53.6 
                & 74.61   & 71.81   \\ 
\textbf{mGPT-2 (gen)}  & 69.78   & 67.57  
                & 45.6   & 43.48   
                & 63.64   & 58.75   \\ 
\textbf{mGPT-2 (gen\_orig)}  & 58.64   & 62.22   
                & 43.15   & 42.3
                & 50.61   & 52.64   \\ 
\midrule
\textbf{mT5 (NSP)}     & 84.62   & 81.96 
                & 58.08   & 54.83   
                & 70.95   & 74.05  \\ 
\textbf{mT5 (gen)}     & 31.56   & 32.6
                & 52.85   & 53.93 
                & 29.76   & 30.03   \\

\bottomrule 
\end{tabular}
}
\end{table}

%----------------------------------------------------------------------------------------

\section{Score Calculation Differences}
\label{a:diff_nadeem}

For calculating the LMS, the code published by \cite{nadeem-etal-2021-stereoset} contradicts the explanation in the paper to some extent. In the paper, the calculation is written to be counted towards the meaningful example for "either stereotypical or anti-stereotypical" candidate's superiority; indeed, it is counted towards "both stereotypical and anti-stereotypical" candidate's superiority. The difference would be apparent in an example where the stereotypical candidate's probability is higher than the unrelated candidate's, which is in turn higher than the anti-stereotypical candidate's. In this example, the score would be 100\% according to the paper; however, it would be 50\% according to the code that the authors published. Our approach is based on the code that they published since the same results with their publication are indeed reached by the code.

%----------------------------------------------------------------------------------------

\onecolumn

\section{Multi-Class Results for Intra- and Inter-Sentence Tests}
\label{a:mc_eval}

\begin{table}[ht]
\centering
\caption{Multi-class evaluation results for \textit{intra}-sentence tests on monolingual (top) and multilingual (bottom) models for each language.}
\label{table:mc_intrasentence_eval}
\resizebox{1\textwidth}{!}{
\begin{tabular}{cccccccccccccccc} 
\toprule
 & \multicolumn{5}{c}{\textbf{LMS}} & \multicolumn{5}{c}{\textbf{SS}} & \multicolumn{5}{c}{\textbf{ICAT (Macro / Micro)}}  \\ 

& en & de & tur & fr & es & en & de & tur & fr & es & en & de & tur & fr & es\\
\cmidrule(l{4pt}r{4pt}){2-6} \cmidrule(l{4pt}r{4pt}){7-11} \cmidrule(l{4pt}r{4pt}){12-16}
\textbf{BERT}   & 83.02   & 71.78   & 69.11   & 50.28   & 76.28 
                & 58.63   & 55.28   & 50.87   & 47.83   & 56.2 
                & (64.57/68.69)   & (57.35/64.2)   & (56.6/67.91)   & (41.28/48.1)  & (62.26/66.83) \\ 
\textbf{GPT-2}  & 91.11   & 79.8   & 73.43   & 80.01   & 79.06 
                & 61.93   & 58.42   & 53.23   & 59.66   & 58.65 
                & (66.69/69.37)   & (61.8/66.37)   & (60.76/68.68)   & (59.42/64.54)  & (60.64/65.38) \\ 
\textbf{T5}     & 79.04   & 67.62   & ---     & 50.67   & 63.32 
                & 59.98   & 55.27   & ---     & 54.02   & 55.27 
                & (60.03/63.26)   & (53.47/60.49)   & ---   & (41.16/46.6)  & (49.61/56.65) \\ 
\midrule
\textbf{mBERT}  & 69.94   & 65.67   & 59.28   & 62.09   & 60.05 
                & 52.36   & 49.13   & 50.18   & 52.6   & 51.97 
                & (56.58/66.64)   & (53.59/64.5)   & (46.26/59.06)   & (49.21/58.85)  & (48.53/57.68) \\ 
\textbf{mGPT-2} & 86.52   & 77.12   & 71.44   & 66.89   & 70.74 
                & 55.18   & 50.13   & 52.76   & 48.65   & 55.16 
                & (69.18/77.56)   & (65.06/76.91)   & (60.43/67.5)   & (54.05/65.09)  & (56.3/63.43) \\ 
\textbf{mT5}    & 69.97   & 73.99   & 55.66   & 55.57   & 56.93 
                & 52.56   & 54.19   & 51.41   & 50.85   & 53.97 
                & (55.69/66.39)   & (59.63/67.8)   & (45.8/54.09)   & (46.47/54.62)  & (47.77/52.41) \\ 
\bottomrule 
\end{tabular}
}
\end{table}

\begin{table}[ht]
\centering
\caption{Multi-class evaluation results for \textit{inter}-sentence tests on monolingual (top) and multilingual (bottom) models for each language.}
\label{table:mc_intersentence_eval}
\resizebox{1\textwidth}{!}{
\begin{tabular}{cccccccccccccccc} 
\toprule
 & \multicolumn{5}{c}{\textbf{LMS}} & \multicolumn{5}{c}{\textbf{SS}} & \multicolumn{5}{c}{\textbf{ICAT (Macro / Micro)}}  \\ 

& en & de & tur & fr & es & en & de & tur & fr & es & en & de & tur & fr & es\\
\cmidrule(l{4pt}r{4pt}){2-6} \cmidrule(l{4pt}r{4pt}){7-11} \cmidrule(l{4pt}r{4pt}){12-16}
\textbf{BERT}   & 88.53   & 79.76   & 83.73   & 61.08   & 41.78 
                & 60.43   & 55.8   & 54.41   & 43.72   & 49.22 
                & (67.13/70.06)   & (65.48/70.5)   & (66.72/76.35)   & (49.32/53.41)  & (35.33/41.13) \\  
\textbf{GPT-2}  & 76.37   & 77.04   & 66.42   & 66.5   & 66.96 
                & 52.17   & 51.79   & 49.68   & 50.33   & 47.24 
                & (65.46/73.06)   & (64.43/74.28)   & (56.25/65.99)   & (55.49/66.06)  & (57.61/63.25) \\ 
\textbf{T5}     & 88.59   & 84.55   & ---     & 80.98   & 76.99 
                & 60.71   & 57.44   & ---     & 56.46   & 55.26 
                & (67.36/69.61)   & (67.99/71.97)   & ---   & (64.4/70.51)  & (63.73/68.88) \\ 
\midrule
\textbf{mBERT}  & 83.06   & 77.4   & 78.75   & 77.68   & 76.76 
                & 58.1   & 57.99   & 53.77   & 57.23   & 57.71 
                & (65/69.61)   & (59.95/65.02)   & (64.39/72.81)   & (62.2/66.44)  & (60.27/64.92) \\ 
\textbf{mGPT-2} & 69.83   & 67.43   & 63.74   & 68.52   & 67.45 
                & 45.92   & 43.64   & 48.1   & 45.24   & 45.01 
                & (58.17/64.12)   & (54.34/58.85)   & (53.17/61.32)   & (55.63/62)  & (53.59/60.72) \\ 
\textbf{mT5}    & 84.59   & 81.92   & 79.03   & 82.34   & 82.9 
                & 58.18   & 54.99   & 52.62   & 54.97   & 56.99 
                & (66.31/70.75)   & (64.98/73.75)   & (64.85/74.89)   & (68.2/74.16)  & (64.57/71.31) \\ 
\bottomrule 
\end{tabular}
}
\end{table}

%----------------------------------------------------------------------------------------

\end{document}